\documentclass[letterpaper, 10 pt, conference]{ieeeconf}  

\IEEEoverridecommandlockouts                              

\usepackage{booktabs}
\usepackage{multirow}
\usepackage{graphicx}
\usepackage{comment}
\usepackage{amsfonts}
\usepackage{stfloats}
\usepackage{amsmath}
\usepackage{tabularx} 
\usepackage{url}
\usepackage[colorlinks=true, linkcolor=black, citecolor=black, urlcolor=blue]{hyperref}

\title{Controlling Intent Expressiveness in Robot Motion with Diffusion Models}

\author{
  Wenli Shi, Clemence Grislain, Olivier Sigaud, Mohamed Chetouani \\
  Sorbonne Université, CNRS, Institut des Systèmes Intelligents et de Robotique (ISIR) \\
  Paris, France\\
  \texttt{\{wenli, grislain, sigaud, chetouani\}@isir.upmc.fr} \\
}

\begin{document}
\maketitle

\begin{abstract}

Legibility of robot motion is critical in human–robot interaction, as it allows humans to quickly infer a robot’s intended goal. Although traditional trajectory generation methods typically prioritize efficiency, they often fail to make the robot’s intentions clear to humans. Meanwhile, existing approaches to legible motion usually produce only a single “most legible” trajectory, overlooking the need to modulate intent expressiveness in different contexts. In this work, we propose a novel motion generation framework that enables controllable legibility across the full spectrum—from highly legible to highly ambiguous motions. We introduce a  modeling approach based on an Information Potential Field to assign continuous legibility scores to trajectories, and build upon it with a two-stage diffusion framework that first generates paths at specified legibility levels and then translates them into executable robot actions. Experiments in both 2D and 3D reaching tasks demonstrate that our approach produces diverse and controllable motions with varying degrees of legibility, while achieving performance comparable to SOTA. Code and project page: \url{https://legibility-modulator.github.io/}.

\end{abstract}

\section{Introduction}

When collaborating with humans, robots must not only reach their goals efficiently but also move in ways that make their intentions clear. A trajectory that minimizes path length may be efficient, but when multiple possible goals exist, it can appear ambiguous to an observer. In contrast, a legible trajectory enables human observers to quickly and confidently infer the true goal of the robot~\cite{dragan2013legibility}. Thus, legibility is critical for fluent and safe interaction.

Various methods have been proposed to generate legible robot motion, ranging from optimization-based planners~\cite{amirian2024legibot, luo2024potential} to learning-driven generative approaches~\cite{bronars2024legibility}. These works show that trajectories can be designed to make a robot’s goal more transparent to humans. However, most existing approaches focus on producing a single maximally legible trajectory. This overlooks the fact that legibility is not binary, but a spectrum: in cooperative tasks, robots may need to be highly transparent; in time-critical tasks, they may prioritize efficiency; and in adversarial contexts, they may even benefit from intentional ambiguity. Current methods do not provide a principled mechanism for modulating this degree of intent expressiveness. 

To address this need, we propose a framework for generating trajectories with controllable levels of legibility. Our key idea is to introduce an Information Potential Field (IPF) that models the probabilistic structure of goal inference and yields a potential-aware legibility score. This score enables smooth modulation across the legibility spectrum.

\begin{figure}[!t]
    \centering
    \includegraphics[width=1.\linewidth]{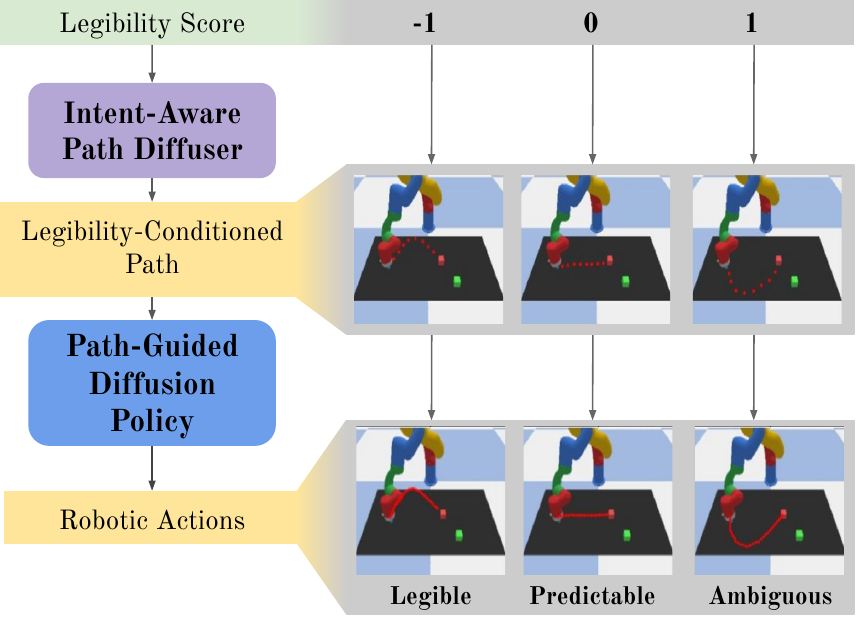}
    \caption{\textbf{Overview of Legibility Modulator.} 
The \textit{Intent-Aware Path Diffuser} generates an intermediate path conditioned on the desired intent expressiveness (legibility score). The \emph{Path-Guided Diffusion Policy} then translates this path into executable robot actions, producing motion that aligns with the specified level of legibility.}
    \label{fig:architecture}
\vspace{-0.6cm}
\end{figure}

To support this approach, we first construct a diverse trajectory dataset inspired by the Quality-Diversity paradigm~\cite{mouret2015mapping}. This dataset covers a wide range of deviation patterns that are retrospectively labeled with potential-based legibility scores, ensuring broad coverage of the legibility space. Building on this foundation, we propose \textbf{Legibility Modulator}, a two-stage diffusion framework illustrated in Figure~\ref{fig:architecture}.

We validate our method in 2D and 3D reaching tasks with various goal configurations. Results show that Legibility Modulator produces diverse and controllable trajectories across the entire legibility spectrum, while maintaining task success and matching the performance of state-of-the-art baselines in maximally legible settings.

Our contributions are threefold:

\begin{enumerate}
    \item We introduce a new legibility-score based on an information potential field, enabling continuous scoring of trajectories in terms of intent expressiveness.  
    \item We construct a trajectory dataset using a quality-diversity paradigm to generate diverse trajectories across the legibility spectrum.
    
    \item We propose a two-stage diffusion approach trained on this dataset, achieving performance comparable to SOTA methods while additionally enabling fine-grained control of intent expressiveness from ambiguous to highly legible behaviors.
\end{enumerate}

\section{Related Work}

{\bf Legible Robot Motion:}
Legible motions enable human observers to quickly and confidently infer the intended goal of a robot, which in most studies corresponds to a spatial target in the workspace. Dragan et al.~\cite{dragan2013legibility} provided a probabilistic formulation distinguishing legibility from predictability, highlighting its importance for safe and fluent human–robot interaction~\cite{chakraborti2019explicability}. Early works employed optimization-based planners that shaped cost functions on trajectory characteristics such as distance to alternative goals~\cite{bodden2018flexible}. Although effective, these approaches required hand-crafted heuristics and lacked scalability. Learning-based methods have since emerged~\cite{zhao2020actor, busch2017learning}, including reinforcement learning to optimize legibility metrics~\cite{bied2020integrating} and supervised observer models to predict human goal inference~\cite{wallkotter2022slot}.

{\bf Robot Motion from Conditional Generative Models:}
Conditional generative models have recently shown strong capabilities in robot motion generation~\cite{wolf2025diffusion}. They naturally capture multi-modality and provide stable training. In particular, diffusion models~\cite{ho2020denoising} have been applied to both action- and trajectory-level generation~\cite{chi2024diffusionpolicy, bronars2024legibility}, offering temporal consistency and robustness to multi-modality. Methods such as classifier-free guidance~\cite{ho2022classifier} further allow the sampling process to be biased toward conditional distributions in a controllable manner. Building on these, \cite{bronars2024legibility} proposed Legibility Diffuser, a diffusion-based policy that leverages guided sampling to imitate the most legible modes from demonstrations. However, such approaches focus on producing a single maximally legible trajectory. Since legibility is inherently a spectrum that must be modulated depending on the context, this limitation motivates our work.

{\bf Potential Fields:}
Artificial Potential Fields (APF), introduced by Khatib~\cite{khatib1986real}, have been widely used for navigation and obstacle avoidance. Beyond basic planning, they have also been combined with generative models. Luo et al.~\cite{luo2024potential} proposed potential-based diffusion planning to avoid local minima, while 
\cite{schmidt2024through} applied entropy-scaled fields to improve legibility in cluttered environments. However, prior work treats potential fields as geometric constructs that ensure feasibility, without accounting for the probabilistic nature of human goal inference. In contrast, our work extends this paradigm by introducing a probabilistic potential formulation tailored to legibility-aware motion generation.

\section{Problem Statement}

We formalize the problem of legible motion generation as a goal-conditioned Markov Decision Process (MDP), defined as: $
\mathcal{M} = (\mathcal{S}, \mathcal{A}, \mathcal{T}, \rho_0, \mathcal{G}),
$ where $\mathcal{S}$ is the state space, typically the robot's configuration or the pose of the end effector, $\mathcal{A}$ is the action space,  
 $\mathcal{T}(s'|s,a)$ is the transition function,  
 $\rho_0$ is the initial state distribution and
$\mathcal{G}$ is the finite set of target positions.
In this framework, we define a \textit{trajectory} as a sequence of state–action pairs $\xi = (s_0, a_0, s_1, a_1, \dots, s_T)$, and a \textit{path}, a subsampled sequence of states $\tau = (s_n, s_{2n}, \dots, s_{kn})$ where $n$ is the step size controlling the temporal resolution.
Given a start state $s_0$ and an intended goal $g^* \in \mathcal{G}$, the robot must generate a trajectory $\xi$ that reaches $g^*$ while maximizing the human observer's ability to infer this goal early and unambiguously, distinguishing it from alternative non-goals $g^- \in \mathcal{G}\setminus\{g^*\}$.

\section{Predictability and Legibility}

In multi-goal settings, robot trajectories must balance efficiency with intent expressiveness. Two key but distinct notions are \textit{predictability} and \textit{legibility}~\cite{dragan2013legibility, chakraborti2019explicability}.

\textbf{Predictability.} Given a known goal $g$, predictability measures how well a trajectory aligns with an observer's expectation, typically some minimal cost $C(\xi)$ (e.g., path length). In other words, the most predictable path is simply the lowest-cost trajectory from the start to $g$.

\textbf{Legibility.} Legibility reverses this reasoning: from a partial trajectory $\xi_{s \to q}$, the observer infers the goal that maximizes the posterior probability $P(g|\xi_{s \to q})$. A trajectory is more legible if this probability mass quickly concentrates on the true goal $g^*$. More generally, the legibility of $\xi$ can be quantified as~\cite{dragan2013legibility}:
\begin{equation}\label{eq_legi}
    \text{Legibility}(\xi) = \frac{\int_0^T P(g^*|\xi_{[0 \rightarrow t]}) f(t)\, dt}{\int_0^T f(t)\, dt},
\end{equation}
where $f(t)$ emphasizes early cues and $T$ is the time horizon.

In a reaching task with two adjacent targets, a straight line is efficient and thus predictable but ambiguous, whereas an early deviating curve is less efficient yet more legible.




\section{Methods}

\begin{figure*}[!ht]
    \centering
    \includegraphics[width=1.\textwidth]{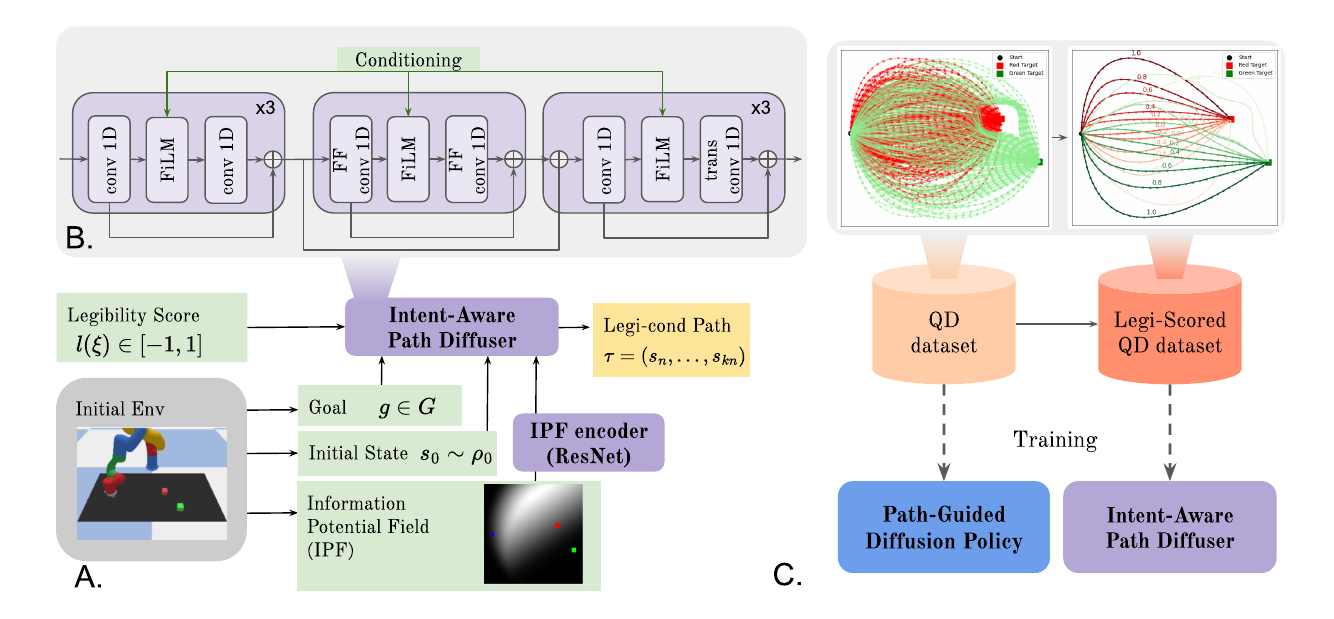}
    \caption{\small \textbf{Detailed architecture of our two-stage conditional diffusion framework.} 
(A) The framework takes as input the initial environment state $s_0$, a candidate goal $g$, the IPF that encodes the probabilistic structure of goal inference, and a continuous \emph{legibility score} $\ell(\xi)$ specifying the desired level of expressiveness. (B) The Intent-Aware Path Diffuser is a U-Net conditioned via FiLM layers on the inputs, which generates a path $\tau$ using diffusion process. (C) The Path-Guided Diffusion Policy is trained on trajectories generated by the QD algorithm, using the corresponding paths as conditioning inputs. The Intent-Aware Path Diffuser is trained on the same QD trajectories, labeled with potential-based legibility scores.
}
    \label{fig:main_fig}
    \vspace{-0.5cm}
\end{figure*}

Our goal is to generate robot motions that are both goal-directed and legible to human observers, with controllable levels of legibility. 
To this end, we propose \textbf{Legibility Modulator}, illustrated in Figure~\ref{fig:main_fig}. 

\subsection{Information Potential Field for Legibility Modeling}

In the formal definition of legibility (Eq.~\eqref{eq_legi}), the posterior $P(g^*|\xi_{[0 \rightarrow t]})$ reflects how observers infer goals from motion prefixes, but is impractical to obtain directly.  
We therefore approximate it with an \emph{Information Potential Field} (IPF), which considers spatial configurations but omits dynamic aspects like velocity.

The likelihood of observing configuration $x$ when aiming for $g$ is modeled as $P(x|g) \sim \mathcal{N}(x_g, \sigma^2 I)$.

The posterior over the intended goal $g^*$ is then
\begin{equation}
    P(g^*|x) = \frac{P(x|g^*)}{P(x|g^*) + \sum_{g^- \in \mathcal{G}\setminus\{g^*\}} P(x|g^-)}.
\end{equation}
This posterior naturally encodes the competition between the intended goal and opposing ones: evidence for $g^*$ increases $P(g^*|x)$, while support for distractors reduces it. 

We define the \emph{information potential} as the negative log-posterior probability of goal $g^*$ given observation $x$:
\begin{equation}
    \phi(x|g^*) = -\log P(g^*|x),
\end{equation}
which is small when $x$ supports $g^*$ and large when ambiguous. This potential provides a continuous, tractable surrogate for goal disambiguation along a trajectory.
For practical representation, the IPF can be stored as a spatial grid: in 2D tasks as a single-channel image $1 \times H \times W$, and in 3D tasks as a volumetric grid $D \times H \times W$.

\subsection{Trajectory Dataset Generation}

To enable trajectory generation across the full spectrum of legibility, we construct a dataset following the quality-diversity (QD) paradigm~\cite{mouret2015mapping,huber2024domain}.  
The goal is to obtain trajectories that are both feasible and well spread in terms of their intent expressiveness, covering legible, predictable, and ambiguous behaviors.

Each trajectory is parameterized as a cubic Bézier curve~\cite{hwang2003mobile} between the start $s$ and goal $g^*$, with control points $c_1,c_2$ sampled within bounds. Curves are discretized by arc-length to ensure consistent resolution.  
To characterize diversity, we extract the \emph{most deviating point} from the straight line $\overline{sg^*}$ and use its spatial position as a descriptor. By uniformly partitioning the workspace into grid cells based on this deviation point, we obtain a set of trajectories that cover different geometric styles and legibility levels.

Each discretized trajectory $\xi=\{x_t\}_{t=1}^T$ is further assigned a potential-based legibility score:
\begin{equation}\label{eq-lp}
    L_p(\xi) = -\sum_{t=1}^T f(t)\, \phi(x|g^*), \quad f(t)=\exp(-\alpha t),
\end{equation}
where earlier points are weighted more strongly. Larger $L_p(\xi)$ corresponds to higher legibility.  
We normalize the scores into $[-1,1]$ by ranking within each target set using, $\ell(\xi) = 2 \cdot \frac{\text{rank}(L_p(\xi))}{N} - 1$, 
where $\ell(\xi)=1$ denotes maximally legible, $\ell(\xi)=-1$ maximally ambiguous, and $\ell(\xi)=0$ maximally predictable trajectories.
This procedure yields a dataset that is both geometrically diverse and uniformly distributed along the legibility spectrum, forming a principled basis for conditional generation.

\subsection{Two-Stage Conditional Diffusion Framework}

\noindent
{\bf Stage 1: Intent-aware Path Diffuser:}
Since legibility depends on global trajectory shape rather than low-level actions, the first stage generates legibility-conditioned paths $\tau=\{s_n,\dots,s_kn\}$ (we set $k=8$ for compactness).  
The diffusion model is conditioned on three signals: (i) start and goal $(s,g)$, (ii) Information Potential Field (IPF) features, and (iii) the legibility label $\ell(\xi)$.  
IPF features are encoded as images in both 2D and 3D environments, enabling integration with convolutional backbones.  
These inputs are fused into a context vector and injected into a UNet denoiser via FiLM~\cite{perez2018film}, which refines Gaussian noise into a legibility-consistent trajectory plan.

\noindent
{\bf Stage 2: Path-Guided Diffusion Policy:}
The second stage converts the path into executable actions.  
It follows the diffusion policy framework~\cite{chi2024diffusionpolicy} and is conditioned on: (i) a short state history, (ii) the goal $g$, and (iii) the path $\tau$.  
The policy outputs an action sequence $(a_1,\dots,a_H)$ that reproduces the spatial plan while satisfying dynamics.  

Both models are trained independently -~Stage 1 on trajectories labeled with legibility scores, and Stage 2 on state–action demonstrations~- and are composed sequentially at inference ensuring control of legibility at the trajectory level and feasibility.

\section{Experiments}
\label{sec-exp}


We compare our \textbf{Legibility Modulator} against three baselines: \textbf{Legibility Diffuser}~\cite{bronars2024legibility}, a diffusion-based policy with guided sampling toward legible modes, \textbf{Diffusion Policy}~\cite{chi2024diffusionpolicy}, a standard conditional diffusion baseline without legibility guidance, and \textbf{Dataset Oracle}, the most legible trajectories from demonstrations, serving as an upper bound.

We evaluate them on the \textit{Block Reaching} task, a standard benchmark for legible motion~\cite{dragan2013legibility,wallkotter2022slot}, where a robot must reach one of two possible blocks. 
We consider two variants:  
(i) a 2D setting modeling end-effector motion in a plane, adapted from prior multi-modal policy tasks~\cite{shafiullah2022behavior,florence2021implicit}, and  
(ii) a 3D setting in PandaGym~\cite{gallouedec2021pandagym} with a Franka arm, closer to real manipulation.
We report two complementary legibility metrics.  
\par\smallskip $\bullet$
\textbf{Distance-based legibility score} used in~\cite{bronars2024legibility} measures geometric divergence from distractor goals:
\begin{equation}\small
    L_d(\xi) = \sum_{t=1}^T \frac{\| g^- - x_t \|^2}{t}
\end{equation}
where $g^-$ is the non-target goal. Higher values indicate clearer separation, though the metric is less sensitive to subtle legibility differences.  

\par\smallskip $\bullet$
\textbf{Potential-based legibility score} uses our IPF measure in Eq.~\eqref{eq-lp} with $f(t)=-1/t$, for a fair comparison with $L_d$. Lower values indicates more legible motions.


\section{Results}


\subsection{Comparison in the Maximally Legible Regime}

Most prior methods on legible motion generation focus on producing a single maximally legible trajectory.  
To align with this setting, we first fix the conditioning label to $\ell(\xi)=1$ and compare against the baselines.
Table~\ref{tab:baseline_compare} shows that our method matches the performance of the Legibility Diffuser on both $L_d$ and $L_p$, while maintaining perfect task success. Furthermore, it achieves significant improvements in $L_d$ and $L_p$ for both 2D and 3D environments compared to the legibility-unaware Diffusion Policy. These results indicate that our framework produces highly legible motion predictions that align with SOTA legibility-aware methods when evaluated under the maximally legible regime established in the literature.


\begin{table}[htbp]
\footnotesize
\caption{
Comparison of our method under the maximally legible regime $\ell(\xi)=1$ with baselines, on the 2D and 3D block reaching environments. In bold are the results that are significantly different from legibility-unaware methods. }
\centering
\begin{tabularx}{\columnwidth}{l *{5}{>{\centering\arraybackslash}X}}
\toprule
& \textbf{SR} & \multicolumn{2}{c}{\textbf{Target R}} & \multicolumn{2}{c}{\textbf{Target G}} \\
\cmidrule(lr){3-4}\cmidrule(lr){5-6}
\textbf{Method (2D Env)} &  & \(L_d \uparrow\) & \(L_p \downarrow\) & \(L_d \uparrow\) & \(L_p \downarrow\) \\
\midrule
Diffusion Policy    & 1. & 1.54 & $1.5\mathrm{e}^{-3}$ & 1.23 & $2.52\mathrm{e}^{-1}$ \\
Legibility Diffuser  & 1. & \textbf{1.97} & $\mathbf{3}\mathrm{e}^{-4}$  & \textbf{1.67} & $\mathbf{4.62}\mathrm{e}^{-2}$ \\
Legi Modulator (Ours)      & 1. & \textbf{1.96} & $\mathbf{4}\mathrm{e}^{-4}$ & \textbf{1.65} & $\mathbf{4.19}\mathrm{e}^{-2}$ \\
\midrule
Oracle (Max Leg Traj) & - & 2.02 & 3$\mathrm{e}^{-4}$ & 1.69 & 3.24$\mathrm{e}^{-2}$ \\
\bottomrule
\toprule
\textbf{Method (3D Env)} &  &  & & &  \\
\midrule
Diffusion Policy    & 1. & 3.05 & 3.01$\mathrm{e}^{-1}$ & 3.24 & 2.80$\mathrm{e}^{-2}$ \\
Legibility Diffuser  & 1. & \textbf{3.85} & \textbf{3.05}$\mathrm{e}^{-2}$ & \textbf{3.95} & \textbf{3.30}$\mathrm{e}^{-3}$ \\
Legi Modulator (Ours)   & 1. &\textbf{3.88} & \textbf{2.18}$\mathrm{e}^{-2}$ & \textbf{3.95} & \textbf{1.50}$\mathrm{e}^{-3}$  \\
\midrule
Oracle (Max Leg Traj) & - & 3.94 & 1.24$\mathrm{e}^{-2}$ & 3.74 & 3.10$\mathrm{e}^{-3}$ \\
\bottomrule
\end{tabularx}
\label{tab:baseline_compare}
\vspace{-0.5cm}
\end{table}

\subsection{Controllable Legibility Spectrum}

\begin{figure}[htbp]
    \centering
    \includegraphics[width=0.99\linewidth]{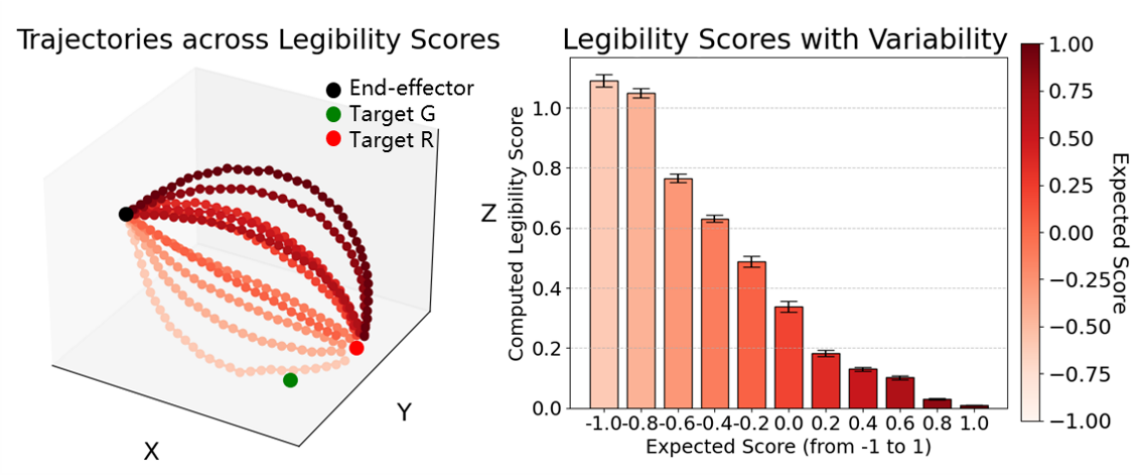}
    \caption{Results in the 3D Block Reaching (Target Red), showing smooth modulation from ambiguous to legible motions. Similar behaviors are also observed for the alternative target and in the 2D environment.}
    \label{fig:result}
\end{figure}

Beyond the maximally legible setting, Legibility Modulator uniquely enables fine-grained control of legibility. By varying the conditioning label $\ell(\xi)$ from $-1$ (most ambiguous) through $0$ (predictable) to $1$ (most legible), trajectories evolve smoothly and qualitatively align with the intended expressiveness. 
Figure~\ref{fig:result} illustrates this effect in the 3D reaching task. On the left, we show example trajectories generated by our method, colored from light to dark red to indicate increasing expected legibility scores. When conditioned on a low legibility score (lighter red), Legibility Modulator produces ambiguous trajectories with respect to the true goal (red). In contrast, when conditioned on a high legibility score (darker red), it produces highly legible trajectories from which a human observer can better infer the intended goal. On the right, we show that the potential-based score $L_p$ decreases smoothly and monotonically as $\ell(\xi)$ increases, demonstrating that our method achieves fine-grained modulation across the legibility spectrum.

\noindent
In summary, Legibility Modulator not only matches existing methods under maximally legible settings but also provides a principled mechanism for modulating intent expressiveness across the full spectrum, while preserving task success.

\section{Conclusion}

We presented Legibility Modulator, a two-stage diffusion framework for generating robot motion with controllable levels of legibility, combining an Information Potential Field for trajectory scoring with an intent-aware path diffuser and a path-guided policy. Experiments in 2D and 3D reaching tasks showed that our approach not only matches SOTA performance in maximally legible settings but also enables smooth modulation across the entire legibility spectrum. A limitation of this study is that experiments were restricted to simulated reaching tasks, which do not fully capture the complexity of real human–robot interaction. In future work, we plan to extend our framework to more naturalistic settings and evaluate its effectiveness with human participants.  

\section*{Acknowledgments}
This work was performed using HPC resources from GENCI–IDRIS (Grant 2025-[AD011016424]). This research was partially funded by the French National Research Agency (ANR) under the OSTENSIVE project “ANR-24-CE33-6907-01.”

\bibliographystyle{IEEEtran}
\bibliography{refs}

\begin{thebibliography}{10}
\providecommand{\url}[1]{#1}
\csname url@rmstyle\endcsname
\providecommand{\newblock}{\relax}
\providecommand{\bibinfo}[2]{#2}
\providecommand\BIBentrySTDinterwordspacing{\spaceskip=0pt\relax}
\providecommand\BIBentryALTinterwordstretchfactor{4}
\providecommand\BIBentryALTinterwordspacing{\spaceskip=\fontdimen2\font plus
\BIBentryALTinterwordstretchfactor\fontdimen3\font minus \fontdimen4\font\relax}
\providecommand\BIBforeignlanguage[2]{{%
\expandafter\ifx\csname l@#1\endcsname\relax
\typeout{** WARNING: IEEEtran.bst: No hyphenation pattern has been}%
\typeout{** loaded for the language `#1'. Using the pattern for}%
\typeout{** the default language instead.}%
\else
\language=\csname l@#1\endcsname
\fi
#2}}

\bibitem{dragan2013legibility}
A.~D. Dragan, K.~C. Lee, and S.~S. Srinivasa, ``Legibility and predictability of robot motion,'' in \emph{2013 8th ACM/IEEE International Conference on Human-Robot Interaction (HRI)}.\hskip 1em plus 0.5em minus 0.4em\relax IEEE, 2013, pp. 301--308.

\bibitem{amirian2024legibot}
J.~Amirian, M.~Abrini, and M.~Chetouani, ``Legibot: Generating legible motions for service robots using cost-based local planners,'' in \emph{2024 33rd IEEE International Conference on Robot and Human Interactive Communication (ROMAN)}.\hskip 1em plus 0.5em minus 0.4em\relax IEEE, 2024, pp. 461--468.

\bibitem{luo2024potential}
Y.~Luo, C.~Sun, J.~B. Tenenbaum, and Y.~Du, ``Potential based diffusion motion planning,'' \emph{arXiv preprint arXiv:2407.06169}, 2024.

\bibitem{bronars2024legibility}
M.~Bronars, S.~Cheng, and D.~Xu, ``Legibility diffuser: Offline imitation for intent expressive motion,'' \emph{IEEE Robotics and Automation Letters}, 2024.

\bibitem{mouret2015mapping}
\BIBentryALTinterwordspacing
J.-B. Mouret and J.~Clune, ``Illuminating search spaces by mapping elites,'' 2015. [Online]. Available: \url{https://arxiv.org/abs/1504.04909}
\BIBentrySTDinterwordspacing

\bibitem{chakraborti2019explicability}
T.~Chakraborti, A.~Kulkarni, S.~Sreedharan, D.~E. Smith, and S.~Kambhampati, ``Explicability? legibility? predictability? transparency? privacy? security? the emerging landscape of interpretable agent behavior,'' in \emph{Proceedings of the international conference on automated planning and scheduling}, vol.~29, 2019, pp. 86--96.

\bibitem{bodden2018flexible}
C.~Bodden, D.~Rakita, B.~Mutlu, and M.~Gleicher, ``A flexible optimization-based method for synthesizing intent-expressive robot arm motion,'' \emph{The International Journal of Robotics Research}, vol.~37, no.~11, pp. 1376--1394, 2018.

\bibitem{zhao2020actor}
X.~Zhao, T.~Fan, D.~Wang, Z.~Hu, T.~Han, and J.~Pan, ``An actor-critic approach for legible robot motion planner,'' in \emph{2020 IEEE international conference on robotics and automation (ICRA)}.\hskip 1em plus 0.5em minus 0.4em\relax IEEE, 2020, pp. 5949--5955.

\bibitem{busch2017learning}
B.~Busch, J.~Grizou, M.~Lopes, and F.~Stulp, ``Learning legible motion from human--robot interactions,'' \emph{International Journal of Social Robotics}, vol.~9, no.~5, pp. 765--779, 2017.

\bibitem{bied2020integrating}
M.~Bied and M.~Chetouani, ``Integrating an observer in interactive reinforcement learning to learn legible trajectories,'' in \emph{2020 29th IEEE International Conference on Robot and Human Interactive Communication (RO-MAN)}.\hskip 1em plus 0.5em minus 0.4em\relax IEEE, 2020, pp. 760--767.

\bibitem{wallkotter2022slot}
S.~Wallk{\"o}tter, M.~Chetouani, and G.~Castellano, ``Slot-v: supervised learning of observer models for legible robot motion planning in manipulation,'' in \emph{2022 31st IEEE International Conference on Robot and Human Interactive Communication (RO-MAN)}.\hskip 1em plus 0.5em minus 0.4em\relax IEEE, 2022, pp. 1421--1428.

\bibitem{wolf2025diffusion}
R.~Wolf, Y.~Shi, S.~Liu, and R.~Rayyes, ``Diffusion models for robotic manipulation: A survey,'' \emph{arXiv preprint arXiv:2504.08438}, 2025.

\bibitem{ho2020denoising}
J.~Ho, A.~Jain, and P.~Abbeel, ``Denoising diffusion probabilistic models,'' \emph{Advances in neural information processing systems}, vol.~33, pp. 6840--6851, 2020.

\bibitem{chi2024diffusionpolicy}
C.~Chi, Z.~Xu, S.~Feng, E.~Cousineau, Y.~Du, B.~Burchfiel, R.~Tedrake, and S.~Song, ``Diffusion policy: Visuomotor policy learning via action diffusion,'' \emph{The International Journal of Robotics Research}, 2024.

\bibitem{ho2022classifier}
J.~Ho and T.~Salimans, ``Classifier-free diffusion guidance,'' \emph{arXiv preprint arXiv:2207.12598}, 2022.

\bibitem{khatib1986real}
O.~Khatib, ``Real-time obstacle avoidance for manipulators and mobile robots,'' \emph{The international journal of robotics research}, vol.~5, no.~1, pp. 90--98, 1986.

\bibitem{schmidt2024through}
M.~Schmidt-Wolf, T.~Becker, D.~Oliva, M.~Nicolescu, and D.~Feil-Seifer, ``Through the clutter: Exploring the impact of complex environments on the legibility of robot motion,'' \emph{arXiv preprint arXiv:2406.00119}, 2024.

\bibitem{huber2024domain}
J.~Huber, F.~H{\'e}l{\'e}non, H.~Watrelot, F.~B. Amar, and S.~Doncieux, ``Domain randomization for sim2real transfer of automatically generated grasping datasets,'' in \emph{2024 IEEE international conference on robotics and automation (ICRA)}.\hskip 1em plus 0.5em minus 0.4em\relax IEEE, 2024, pp. 4112--4118.

\bibitem{hwang2003mobile}
J.-H. Hwang, R.~C. Arkin, and D.-S. Kwon, ``Mobile robots at your fingertip: Bezier curve on-line trajectory generation for supervisory control,'' in \emph{Proceedings 2003 IEEE/RSJ International Conference on Intelligent Robots and Systems (IROS 2003)(Cat. No. 03CH37453)}, vol.~2.\hskip 1em plus 0.5em minus 0.4em\relax IEEE, 2003, pp. 1444--1449.

\bibitem{perez2018film}
E.~Perez, F.~Strub, H.~De~Vries, V.~Dumoulin, and A.~Courville, ``Film: Visual reasoning with a general conditioning layer,'' in \emph{Proceedings of the AAAI conference on artificial intelligence}, vol.~32, no.~1, 2018.

\bibitem{shafiullah2022behavior}
\BIBentryALTinterwordspacing
N.~M.~M. Shafiullah, Z.~J. Cui, A.~Altanzaya, and L.~Pinto, ``Behavior transformers: Cloning $k$ modes with one stone,'' in \emph{Thirty-Sixth Conference on Neural Information Processing Systems}, 2022. [Online]. Available: \url{https://openreview.net/forum?id=agTr-vRQsa}
\BIBentrySTDinterwordspacing

\bibitem{florence2021implicit}
P.~Florence, C.~Lynch, A.~Zeng, O.~Ramirez, A.~Wahid, L.~Downs, A.~Wong, J.~Lee, I.~Mordatch, and J.~Tompson, ``Implicit behavioral cloning,'' \emph{Conference on Robot Learning (CoRL)}, November 2021.

\bibitem{gallouedec2021pandagym}
Q.~Gallou{\'e}dec, N.~Cazin, E.~Dellandr{\'e}a, and L.~Chen, ``{panda-gym: Open-Source Goal-Conditioned Environments for Robotic Learning},'' \emph{4th Robot Learning Workshop: Self-Supervised and Lifelong Learning at NeurIPS}, 2021.

\end{thebibliography}

\end{document}